\documentclass[12pt]{article}
\usepackage{amsmath}
\usepackage{graphicx}
\usepackage{verbatim}
\usepackage{hyperref}
\title{Learning Markov Network Structure using\\ Brownian Distance Covariance}
\author{Ehsan Khoshgnauz\footnote{\href{mailto:ehsankhnz@gmail.com}{\nolinkurl{ehsankhnz@gmail.com}}, 
Department of Statistics, Tehran Payame Noor University, Tehran, Iran.}}

\begin{document}
\maketitle
\begin{abstract}
 In this paper, we present a simple non-parametric method for learning the structure of
undirected graphs from data that drawn from an underlying unknown distribution. 
We propose to use Brownian distance covariance to estimate the conditional independences
between the random variables and encodes pairwise Markov graph. 
This framework can be applied in high-dimensional setting, where the number of parameters much
be larger than the sample size. 
\end{abstract}
\section{Introduction}
Undirected graphical models, also known as Markov random fields or Markov networks, have
become a part of the mainstream of statistical theory and application in recent years. 
These models use graphs to represent conditional independences among sets of random variables. 
In these graphs, the absence of an edge between two vertices means the corresponding random
variables are conditionally independent, given the other variables. 
Learning the structure of a graph is equivalent to learning if there exists
an edge between every pair of nodes in the graph. 
\\
In the past decade, significant progress has been made on designing efficient algorithms
to learn undirected graphs from high-dimensional observational datasets. Most of these
methods are based on either the penalized maximum-likelihood estimation or penalized regression methods. 
Works has focused on the problem of estimating the graph in this high dimensional setting,
which becomes feasible if graph is sparse.

 \cite{ref.glasso} develop an efficient algorithm for computing the estimator with 
excellent theoretical properties using a graphical version of the lasso.
\\
In high dimensional problems normality assumption is the main constraint in methods. 
But we can replace the Gaussian constraint with a semi-parametric Gaussian copula, as discussed in \cite{ref.npn}. 
This method use a semi-parametric Gaussian copula or non-paranormal approach by replacing linear functions
with a set of one-dimensional smooth functions, for high dimensional inference. 
The non-paranormal extends the normal by transforming the variables by smooth functions. 
\\
The Gaussian distribution is almost always used in study for this scope,
because the Gaussian distribution represents at most second-order
relationships, it automatically encodes a pairwise Markov graph. 
The Gaussian distribution has the property that if the
$ijth$ component of inverse covariance matrix is zero, then variables i and
j are conditionally independent, given the other variables.
\\
\\Distance correlation is a measure of dependence between random vectors introduced by \cite{ref.bdc}.
For all distributions with finite first moments distance correlation,
it is zero if and only if the random vectors are independent. 
We use this properties to construct our method.
\\
In this paper, we have discussed the structure learning of Markov graphs with large-dimensional covariance
matrices where the number of variables is not small compared to the sample size.
It is well-known that in such situations the usual estimator, the sample
covariance matrix, may not be invertible. The approach
suggested is to use distance covariance matrix towards the identity this matrix.\\

\section{The proposed method}
 In this paper we are concerned with the task of estimating
the graph structure of a Markov random field over a random vector
$X = (X_{1},X_{2},. . . ,X_{p})$, given n independent and identically distributed samples.
\\
It was shown that the distance covariance is zero if and only if the two vectors
were independent, we use this property.
\\
 The main idea is to create a matrix from each pair of distance correlation.
Then use it to construct an adjacency matrix of conditional independents between each node pair.
\\
The distance dependence statistics in \cite{ref.bdc} are defined as follows. 
For a random sample $(X,Y) = {(X_{k},Y_{k}) : k = 1,. . . , n}$ of n i.i.d. random vectors (X, Y ) from the
joint distribution of random vectors X in $\Re^{p}$ and Y in $\Re^{q}$, compute the Euclidean
distance matrices $(a_{kl}) = (|X_{k} - X_{l} |_{p})$ and $(b_{kl}) = (|Y_{k} - Y_{l} |_{q} )$.\\
 Define $A_{kl} = a_{kl}-\bar{a}_{k\cdot} - \bar{a}_{\cdot l } + \bar{a}_{\cdot \cdot }, \;\; \;\; k,l = 1,. . . , n,$
\\
where
\begin{align}
\bar{a}_{k\cdot }=\frac{1}{n}\sum_{l=1}^{n}a_{kl},\; \; 
\bar{a}_{\cdot l}=\frac{1}{n}\sum_{k=1}^{n}a_{kl},\; \;
\bar{a}_{\cdot \cdot }=\frac{1}{n^{2}}\sum_{k,l=1}^{n}a_{kl}. 
\end{align}
Similarly define $B_{kl}= b_{kl}-\bar{b}_{k\cdot} - \bar{b}_{\cdot l } + \bar{b}_{\cdot \cdot }, \;\; \;\; k,l = 1,. . . , n,$
\\
Then sample distance correlation $dcor(X,Y)$ are defined by,
\begin{align}
R^{2}(X,Y)=\begin{cases}
\frac{\nu^{2}(X,Y)}{\sqrt{\nu^{2}(X)\nu^{2}(Y)}},& \nu^{2}(X)\nu^{2}(Y)> 0;\\ 
 0,& \nu^{2}(X)\nu^{2}(Y)= 0. 
\end{cases}
\end{align}
where 
\begin{align}
\nu_{n}^{2}(X,Y)=\frac{1}{n^{2}}\sum_{k,l=1}^{n}A_{kl}B_{kl}
\end{align}
is the distance covariance. 
\\
 Also, distance covariance has simple computing formula, the computations would appear
to be $O(n^{2})$, which can be burdensome for large n. 
\\
This estimator is distribution-free and has a simple explicit formula that is easy
to compute and interpret.
\\
In our study this formula become more easier, because here $p=q=1$, for example
in procedure to form distance correlation matrix, $a_{kl}$,$b_{kl}$ for each vector only one time computed,
furthermore $\sum_{k,l=1}$ is changed to $2\sum_{k<l; k,l=1}$ and requires less computing time.
In Appendix \ref{app1}, we show a function coded in R, that calculate this. 
\\
The distance correlation, is implemented in the R package energy \cite{ref.enr}.
\\

\begin{figure}
\begin{center}
\includegraphics{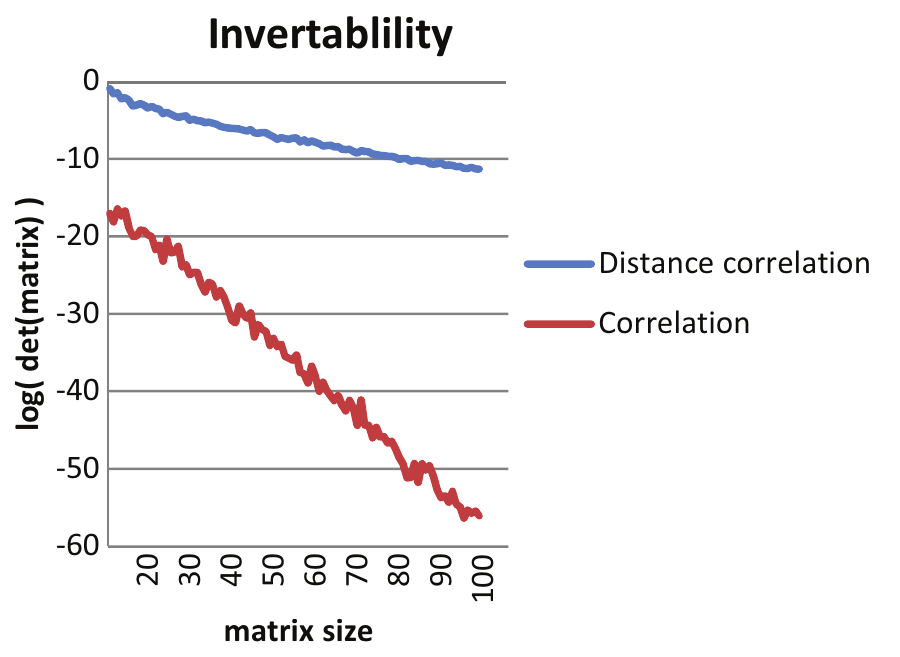} 
\end{center}
\caption{ The logarithm of determinant of correlation matrices in different dimensions.}
\vspace{.3in}
\label{fig4}
\end{figure}

We construct a matrix R of sample distance correlation ($dcor$) between each pair of nodes,
so the element $i,j$ in R is equal to $dcor(X_{i},X_{j})$, 
\\
\begin{align}
\label{dcormatrix}
R=
\begin{pmatrix}
 1& dcor(X_{1},X_{2})  &... &dcor(X_{1},X_{p}) \\ 
 dcor(X_{2},X_{1}) & 1 & ...&dcor(X_{2},X_{p})\\ 
\vdots&\vdots &1&\vdots\\
dcor(X_{p},X_{1}) &dcor(X_{p},X_{2})&...&1
\end{pmatrix}
\end{align}
We call the matrix R defined above the distance correlation matrix.\\
\\
From the property that $dcor(Y,X)=dcor(X,Y)$, obviously R is a symmetric matrix.\\
When the matrix dimension $p$ is larger than the number $n$ of observations available, the ordinary sample
covariance matrix is not invertible. But distance covariance or distance correlation matrix,
has better performance.
\\We generate three different random data in 2 to 100 dimensions, and compute average of determinant of the correlation and 
distance correlation matrix in each dimension.
As we see these results in figure \ref{fig4}, distance correlation is more invertible.
and computationally, is non-singular enough.\\\\

 
Now we construct The partial correlation matrix.
When ${r_{ij}|REST}$ be the partial correlation between
the variables $X_{i}$ and $X_{j}$,
given all the remaining variables, and $P=R^{-1},    P =(p_{{ X_{i}X_{j}}}) $, the inverse of the 
correlation matrix [Whittaker (1990)]; is given by,
\\
 \begin{align}
 \rho_{X_iX_j|REST} = \frac{-p_{ij}}{[p_{ii}p_{jj}]^{\frac{1}{2}}}.
\end{align} 
\\\\\\

We can calculate matrix P, by simple computations.
As in partial.cor function in R Package Rcmdr\cite{ref.rcmdr} implemented, 
following code in R language programming do this.

\begin{verbatim} 
 RI <-solve(R)             #RI is the inverse of R
 D <- 1/sqrt(diag(RI))     
 P <- -RI * (D %o% D)      #%o% is the outer product operator
 diag(P) <- 0
 \end{verbatim}

Finally $P$ is the sparse structure matrix of the graph.\\
We can compare each element of $R$ to a tuning parameter (forming the paths) and derive desired adjacency matrix.

\section{Simulation Results}
\begin{figure}
 \begin{center}
 \includegraphics{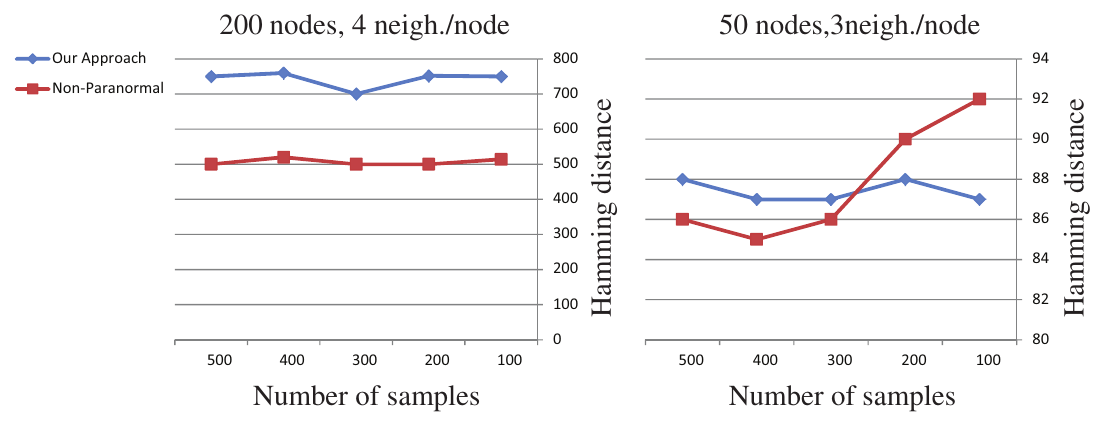} 
 \end{center}
 \caption{The performance comparison between our proposed approach and the non-paranormal approach.} 
 \label{fig5}
\end{figure}

In this simulation, we demonstrate the performance of the
proposed approach on finding the sparse structures of random Markov networks, by generating
Erd\H os-R\'enyi random graphs. \\
The Erd\H os-R\'enyi random graph $Gp$ is a graph on $p$ nodes
in which the probability of an edge being in the graph is $\frac{c}{p}$ and the edges are
generated independently. In this random graph, the average degree of a node is $c$.\\
\\
Given the precision matrix for a
zero-mean Gaussian distribution, it is easy to sample data
from the distribution. But we do not know the distribution.
So we randomly constructed precision matrices, and set random
linear relationships with white noise between the columns of data sample matrices.\\
In a similar manner to \cite{ref.lin}, We simulated Erd\H os-R\'enyi random graphs in two types
of sparse structures (or precision matrices):
1) 50 nodes with averagely 3 neighbours per node, 
2) 200 nodes with averagely 4 neighbours per node.\\
Now, the goal here is to see how well our approach recovers the sparse
structures of those precision matrices given different numbers
of sampled data.\\
Figure \ref{fig5} illustrates the performance of our approach
in recovering the structures of different Markov networks in
comparison with the non-paranormal approach as discussed 
in \cite{ref.npn}.
The performance is evaluated by Hamming distance, the number
of disagreeing edges between an estimated network
and the ground truth, in an equal number of edges.

\section{Experimental Results}
 In this section, we are compared our algorithm to that of non-paranormal method as discussed 
in \cite{ref.npn}, using the function huge.npn() implemented in the R package
huge \cite{ref.huge} for estimating a semi-parametric Gaussian copula model by truncated normal or normal score.
\\

 \begin{figure}
 \begin{center}
 \includegraphics{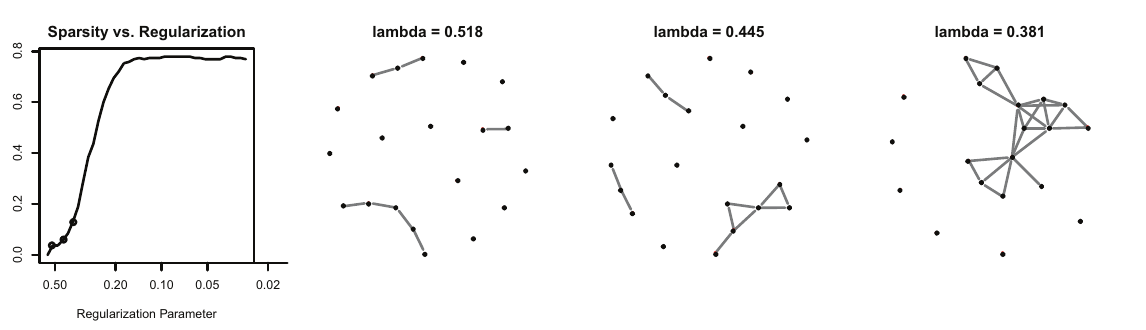} 
 \end{center}
 \caption{ The output of huge, when nlambda=40, lambda.min.ratio =.05 } 
 \vspace{.5in}
 \label{fig1}
\end{figure}

 \begin{figure}
 \begin{center}
 \includegraphics{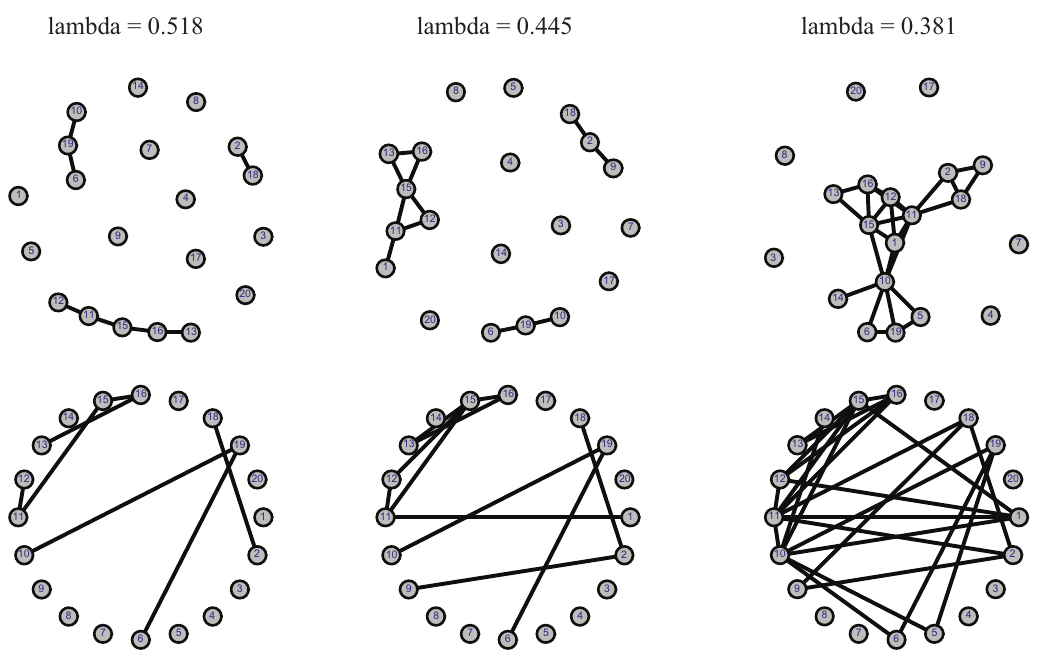} 
 \end{center}
 \caption{The estimated graph paths using non-paranormal method. }
 \vspace{.5in}
 \label{fig2}
\end{figure}

 \begin{figure}
 \begin{center}
 \includegraphics{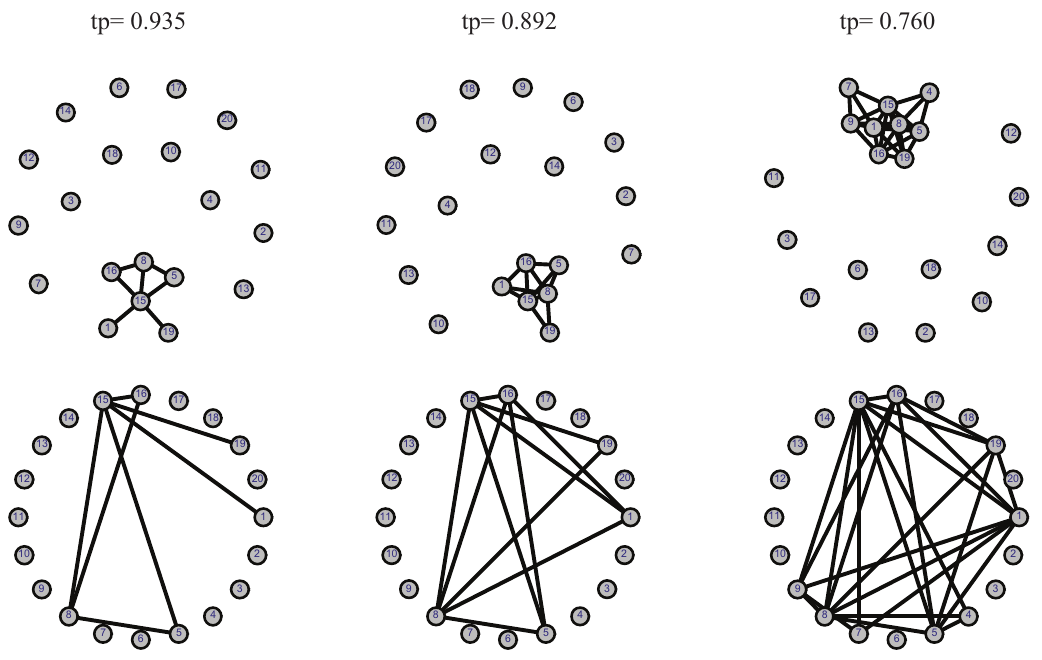} 
 \end{center}
 \caption{The estimated graph paths using our method. Tuning parameter tp, is chosen so that the number of edges is close to non-paranormal method that  plots figure \ref{fig2}.}
 \label{fig3}
\end{figure}

The example is based on a stock market data which is contributed to the huge
package that shows closing prices from all stocks in the $S \& P$ 500 for all days that
the market was open between January 1, 2003 and January 1, 2008. This gave us 1258
samples for the 452 stocks that remained in the $S \& P $ 500 during the entire time period.
Here for convenience and more visibility, we select only first 20 parameters. Also and instead of force-based 
graph drawing layout of Fruchterman-Reingold that utilized in plot function in huge,
we use R package ggm\cite{ref.ggm} for visualize graphs more distinctly. 
\\
The output of huge package graph estimation using the transformed data method and also 
 preprocessing step that mentioned in\cite{ref.huge} is shown in Figure \ref{fig1}.
 \\
Data have been transformed by calculating the log-ratio of the price at time $t$
to price at time $t-1$, and then standardized by subtracting the mean and adjusting the
variance to one. 
\\
 In order to more distinctly, we sending above output estimated data to R package ggm \cite{ref.ggm},
and plot the graph again, in two different layout, circle and Fruchterman-Reingold. Figure \ref{fig2} 
shows the results.
\\

\newpage
\appendix
\section{Appendix}
\label{app1}
\textbf{Simplified distance correlation}\\
The following R code, calculate simplified distance correlation between two vectors x,y.
\begin{small}
\begin{verbatim}
 dcor2<-function(x,y){ 
 n <- length(x);
 if (n != length(y)) {stop("Sample sizes must be equal")}
 u <- matrix(0,2,n+1);
 w<-0
 
for(i in 1:n){
	for(j in 1:n){
	if(i<j){
	w<-abs(x[i]-x[j]);			u[1,i]<-u[1,i]+w;		u[1,j]<-u[1,j]+w;
	}
	else if(i>j){	
	w<-abs(y[i]-y[j]);			u[2,i]<-u[2,i]+w;			u[2,j]<-u[2,j]+w;	
	}		
 } 
 }
 u <-u/n
 u[1,n+1]<-mean(u[1,][1:n])	
 u[2,n+1]<-mean(u[2,][1:n])

 r<-0;	 rx<-0;	 ry<-0;
 for(i in 1:n){ 
	for(j in 1:n){
	r<-  r +(abs(x[i]-x[j])-u[1,i]-u[1,j]+u[1,n+1])*(abs(y[i]-y[j])-u[2,i]-u[2,j]+u[2,n+1])		
	rx<- rx+(abs(x[i]-x[j])-u[1,i]-u[1,j]+u[1,n+1])*(abs(x[i]-x[j])-u[2,i]-u[2,j]+u[2,n+1])		
	ry<- ry+(abs(y[i]-y[j])-u[1,i]-u[1,j]+u[1,n+1])*(abs(y[i]-y[j])-u[2,i]-u[2,j]+u[2,n+1])		
			}
			 }
 rx<-sqrt(rx)/n
 ry<-sqrt(ry)/n
 r<-sqrt(r)/n	
 r/(sqrt(rx*ry)) 
}
\end{verbatim}
\end{small}

\end{document}